\documentclass{article}

\DeclareUnicodeCharacter{0301}{\'{e}}
\usepackage{microtype}
\usepackage{graphicx}
\usepackage{subfigure}
\usepackage{booktabs} 
\usepackage{amsthm}
\usepackage{multirow}
\usepackage{caption}
\usepackage[hidelinks]{hyperref}

\usepackage{hyperref}

\usepackage[accepted]{icml2024}

\usepackage{amsmath}
\usepackage{amssymb}
\usepackage{mathtools}
\usepackage{amsthm}

\usepackage[capitalize,noabbrev]{cleveref}

\theoremstyle{plain}

\theoremstyle{definition}

\theoremstyle{remark}

\usepackage[textsize=tiny]{todonotes}

\icmltitlerunning{Autaptic Synaptic Circuit Enhances Spatio-temporal Predictive Learning of Spiking Neural Networks}

\begin{document}

\twocolumn[
\icmltitle{Autaptic Synaptic Circuit Enhances Spatio-temporal Predictive \\Learning of Spiking Neural Networks}

\begin{icmlauthorlist}
\icmlauthor{Lihao Wang}{pku,xinstitute}
\icmlauthor{Zhaofei Yu}{pku}

\end{icmlauthorlist}

\icmlaffiliation{pku}{Institute for Artificial Intelligence, Peking University, Peking, China.}
\icmlaffiliation{xinstitute}{Shenzhen X-Institute, Shenzhen, China}

\icmlcorrespondingauthor{Zhaofei Yu}{yuzf12@pku.edu.cn}

\icmlkeywords{Machine Learning, ICML}

\vskip 0.3in
]



\printAffiliationsAndNotice{} 

\begin{abstract}

Spiking Neural Networks (SNNs) emulate the integrated-fire-leak mechanism found in biological neurons, offering a compelling combination of biological realism and energy efficiency. In recent years, they have gained considerable research interest. However, existing SNNs predominantly rely on the Leaky Integrate-and-Fire (LIF) model and are primarily suited for simple, static tasks. They lack the ability to effectively model long-term temporal dependencies and facilitate spatial information interaction, which is crucial for tackling complex, dynamic spatio-temporal prediction tasks. To tackle these challenges, this paper draws inspiration from the concept of autaptic synapses in biology and proposes a novel Spatio-Temporal Circuit (STC) model. The STC model integrates two learnable adaptive pathways, enhancing the spiking neurons' temporal memory and spatial coordination. We conduct a theoretical analysis of the dynamic parameters in the STC model, highlighting their contribution in establishing long-term memory and mitigating the issue of gradient vanishing. Through extensive experiments on multiple spatio-temporal prediction datasets, we demonstrate that our model outperforms other adaptive models. Furthermore, our model is compatible with existing spiking neuron models, thereby augmenting their dynamic representations. In essence, our work enriches the specificity and topological complexity of SNNs.

\end{abstract}

\section{Introduction} 

Spiking Neural Networks (SNNs)~\cite{i1} have emerged as a promising paradigm for brain-inspired computation, garnering considerable attention due to their biological plausibility and energy efficiency in comparison to Artificial Neural Networks (ANNs). Similar to real biological neurons, the fundamental computational units in SNNs, known as spiking neurons, exhibit diverse dynamics and possess the ability to process both temporal and spatial information concurrently. Information is encoded and transmitted between spiking neurons through binary spike signals. These spike signals are sparse and event-driven, endowing SNNs deployed on neuromorphic hardware with significant advantages in terms of computational efficiency and energy consumption~\cite{i3,i4,shi2023towards}. Moreover, SNNs offer greater flexibility in integrating relevant biological mechanisms and leveraging insights from neuroscience to address persistent challenges encountered in ANN models, such as catastrophic forgetting~\cite{i5} and robustness~\cite{i6}.

Due to the non-differentiable nature of the spiking process, training SNNs directly through backpropagation has been a challenging task. However, the development of surrogate gradient approaches~\cite{i7,i8} has significantly advanced the training of SNNs. Presently, it is feasible to train SNNs with hundreds of layers~\cite{i9}, achieving performance comparable to state-of-the-art ANNs in tasks like image and speech classification~\cite{i10,i11,i12}. Nonetheless, these tasks primarily focus on static feature recognition. It has been observed that the widely adopted LIF model~\cite{i13}, which serves as a standard spiking neuron model, exhibits limitations in capturing deep spatio-temporal information features and modeling spatio-temporal dependencies. As a result, SNNs face notable constraints when dealing with spatio-temporal prediction tasks.

In contrast, all organisms live in a dynamically changing world where we frequently encounter complex, dynamic, and uncertain tasks. The brain is highly specialized~\cite{i14}, comprising numerous distinct types of neurons that exhibit diverse dynamic behaviors and characteristics~\cite{i16,i17}, along with a variety of connectivity patterns~\cite{i18,i19}. This intricate and complex topology has been demonstrated to allow the brain to respond differently to various stimuli and tasks~\cite{i20,i21}. The crucial factor enabling the brain to excel in addressing dynamic problems lies in its capacity to integrate and analyze information at different levels and scales~\cite{i22,i23}.

In this paper, we propose the Spatio-Temporal Circuit Leaky Integrate-and-Fire (STC-LIF) model, a novel spatio-temporal self-connection model, inspired by biological autaptic synapses to enhance the specificity and topological complexity of SNNs, thereby improving their performance on spatio-temporal tasks. The STC-LIF model incorporates the axon-dendrite circuit and axon-soma circuit, which dynamically regulate the input current and historical information through proportional control. 
By leveraging this self-connection circuit, our model effectively promotes the interaction of spiking neurons in both spatial and temporal dimensions. The main contribution can be summarized as: 

\setlength{\itemsep}{0pt}
\begin{itemize}
\setlength{\itemsep}{0pt}
\setlength{\parsep}{0pt} 
\setlength{\parskip}{0pt}
    \item We conduct a comprehensive analysis of LIF and its variants in spatio-temporal predictive learning, and identify the limitations of modeling long-term dependency and representing spatio-temporal information.
    \item We propose the STC-LIF model, which is inspired by biological autaptic synapses, to enhance the ability to extract spatio-temporal features and model spatio-temporal dependencies.
    \item Experimental results demonstrate the effectiveness of the spatio-temporal self-connected circuit in improving the learning of SNNs. Our method achieves remarkable improvements in various spatio-temporal prediction datasets, surpassing other adaptive neuron models.
    \item The concept of spatio-temporal self-connection can be easily transferred and scaled. We extend the application of this method to Parametric Leaky Integrate-and-Fire (PLIF) and Learnable Multi-Hierarchica (LM-H) models, achieving significant performance improvements in spatio-temporal prediction tasks.
\end{itemize}

\section{Related Work} 
\subsection{Learning of Spiking Neural Networks}

The learning algorithms for SNNs can be divided into three main approaches:
unsupervised learning, ANN-to-SNN conversion, and supervised learning. Unsupervised learning approaches primarily utilize bio-plausible learning rules, such as Hebbian learning~\cite{r1} and Spike-Timing-Dependent Plasticity~\cite{r2}. These optimization methods enable local plasticity learning based on the precise timing of spikes, aligning with the inherent characteristics of SNNs. However, they are typically limited to shallow networks~\cite{r3,r4,r5}. When applied to deep networks, unsupervised learning methods necessitate combination with other techniques like clustering algorithms~\cite{r6}.

The second approach involves converting pre-trained ANNs into SNNs without the need for additional training. Early works~\cite{r7,rueckauer2017conversion} focused on approximating the ReLU function using the firing rate of spiking neurons. While these methods achieved lossless accuracy, they required long simulation time steps. More recent research focuses on reducing conversion error to enhance performance at low latency. These techniques include initial potential compensation~\cite{r9}, revised activation function~\cite{bu2021optimal,deng2021optimal}, and spike calibrating~\cite{hao2023bridging}.

The supervised learning approach based on Backpropagation Through Time (BPTT) draws inspiration from the learning of recurrent neural networks. Due to the non-differentiability of the step function, a surrogate function is often used to approximate the gradient~\cite{i7,i8}. This method
enhances the stability and effectiveness of SNN training, serving as a fundamental technique in SNN research~\cite{r18.2,r18.3,r18.4}. To further enhance the performance of SNN, Wang et al.~\yrcite{r13} and Deng et al.~\yrcite{r14} introduced adaptive gradient error computation to address the gradient mismatch between the surrogate and the original functions. 
Additionally, developing new normalization techniques~\cite{r15,r16,r17,r18.5} and residual connections~\cite{i9} can significantly enhance the convergence speed and stability of deep SNNs. Xiao et al.~\yrcite{r18.6} and Zhu et al.~\yrcite{zhuonline} implemented online training for SNNs, largely reducing the memory cost for GPU training.

\subsection{Adaptive Spiking Neural Models}

The fixed-parameter spiking neuron models commonly used for training and inference impose limitations on the representational capacity of SNNs. Bellec~\yrcite{r19} associated the firing threshold range with the spiking activity of the network during training, improving the dynamics of spiking neurons. To further enhance the adaptability of spiking neurons, researchers introduced learnable membrane leakage parameters and threshold parameters~\cite{r20,plif,r22}. 
GLIF ~\cite{glif} utilized gating factors to control the neuronal dynamical patterns, simulating different biological characteristics to expand the representation space and improve heterogeneity. To mitigate the issues of gradient vanishing and divergence, TC-LIF~\cite{tclif} and LM-H~\cite{lmh} introduced dual membrane potentials and corresponding learnable parameters. These modifications significantly enhanced the adaptability of SNNs. However, despite enriching the representation space and adaptability, the model parameters of these approaches remained fixed during the inference stage. This limitation evidently reduced the generalization capability of SNNs when confronted with dynamically changing data.

\section{Preliminaries} 

\begin{figure*}[t]
\vskip 0.2in
\begin{center}
\centerline{\includegraphics[width=0.9\linewidth]{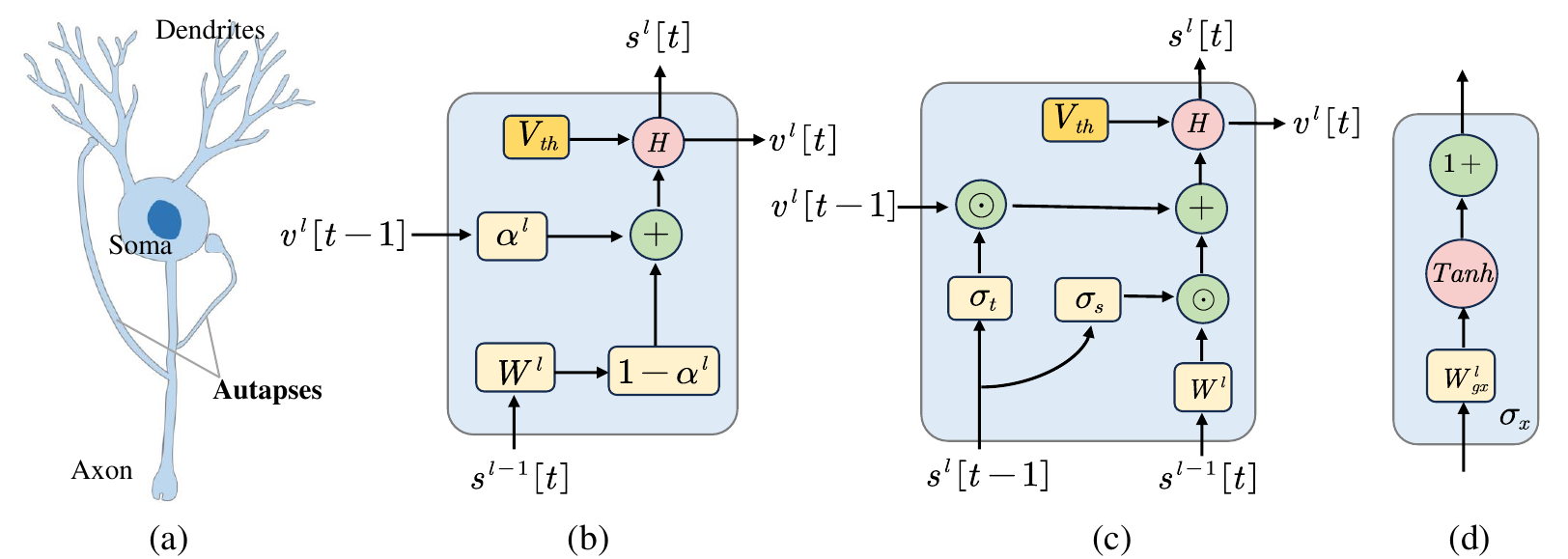}}
\caption{
(a) The structure of Autapses in neuroscience, which includes axon-soma and axon-dendrite circuits. (b) The structure of the vanilla LIF model. (c) The structure of the STC-LIF model. (d) The structure of dynamic spatio-temporal circuit module $\boldsymbol \sigma$ in Figure 1(c).} 
\label{p1}
\end{center}
\vskip -0.2in
\end{figure*}

\subsection{Vanilla LIF Model} 

The vanilla Leaky Integrate-and-Fire (LIF) model is currently the most commonly used spiking neuron model in SNNs. The neural dynamics of the LIF model incorporate three processes of biological neurons~\cite{gerstner2014neuronal}, including integration, firing, and leakage. These processes can be described by the following equations (Figure~\ref{p1}(b)):
\begin{align}
\label{1-1}
\boldsymbol{x}^{l}[t]&=\boldsymbol{W}^l \boldsymbol{s}^{l-1}[t]
\\ 
\boldsymbol{m}^{l}[t]&=\alpha ^l\boldsymbol{v}^{l}[t-1]+( 1-\alpha ^l)\boldsymbol{x}^{l}[t]  \label{1-1-2}
\\
\boldsymbol{s}^{l}[t]&=H (\boldsymbol{m}^{l}[t]-{v}_{th}) \label{1-1-3}
\\
\boldsymbol{v}^{l}[t]&=\boldsymbol{m}^{l}[t]-{v}_{th}\boldsymbol{s}^{l}[t] \label{1-1-4}
\end{align}
Here, $\boldsymbol{x}^{l}[t]$ represents the synaptic current input to layer $l$ at time-step $t$ from presynaptic neurons, and $\boldsymbol{s}^{l}[t]$ represents the spike firing of the neurons in layer $l$ at time-step $t$, the element of which equals 1 if the corresponding neuron fires and 0 otherwise. The parameter $\boldsymbol{W}^l$ denotes the synaptic weight between layer $l$ and the previous layer, $l-1$. $\boldsymbol{m}^{l}[t]$ and $\boldsymbol{v}^{l}[t]$ represent the membrane potential of neurons in layer $l$ before and after firing, respectively. $\alpha^l$is the membrane constant that controls the decay ratio of input and historical information, and it ranges from 0 to 1. When Eq.~\eqref{1-1-2} changes to $\boldsymbol{m}^{l}[t]=\boldsymbol{v}^{l}[t-1]+( 1-\alpha ^l)\boldsymbol{x}^{l}[t]$, the vanilla LIF model degenerates to simple Integrate-and-Fire (IF) model.
$H(\cdot)$ represents the Heaviside step function. Note that this paper employs a soft reset mechanism to keep more information, as suggested by~\cite{rueckauer2017conversion,han2020rmp}. Specifically,
when a spike is fired, the final membrane potential value is reduced by a threshold.

\subsection{Autapses} 
An autapse~\cite{au1} is a unique synaptic structure that refers to a synaptic connection formed between a neuron's axon and its own dendrites or soma (Figure~\ref{p1}(a)). Initially observed in laboratory-cultured neurons, autapses were regarded as erroneous or redundant synaptic connectivity patterns. However, subsequent studies have revealed the presence of autapses in various types of neurons within the central nervous system, including those in the striatum~\cite{au2,au3,au4}, neocortex~\cite{au5}, and hippocampal regions~\cite{au6,au7,au8}. In terms of functionality, inhibitory activation of autapses can suppress the generation of subsequent action potentials~\cite{au9}, thereby regulating the precision of neuronal discharge~\cite{au10}. Conversely, excitatory activation of autapses can enhance neuronal firing, influence neuronal responses, and facilitate coincidence detection~\cite{au11}. These findings highlight the significant regulatory role of autapses in neuronal discharge and emphasize their status as functional structures within the cortical circuits of the brain.

\subsection{Spatio-temporal Predictive Learning} 
Spatio-temporal predictive learning aims to learn future frames based on historical frames. Given a video sequence $X_{t,T}=\left\{ y_i \right\} _{t-T+1}^{t}$ from time $t$ to the past $T$ frames, our objective is to predict the subsequent $T^{\prime}$ frames $Y_{t+1,T^{\prime}}$ $=\left\{ y_i \right\} _{t+1}^{t+T^{\prime}}$, where $y_i\in \mathbb{R} ^{C\times H\times W}$ represents an image with channel number $C$, height $H$, and width $W$. The model with learnable parameters $\varTheta$ learns a mapping $\mathcal{F} _{\varTheta}:X_{t,T}\longmapsto Y_{t+1,T^{\prime}}$, and the optimization objective is:
\begin{align}
\label{3-x}
\varTheta ^*=\mathop {\mathrm{arg}}_{}\mathop {\min}_{\varTheta}\mathcal{L} ( \mathcal{F} _{\varTheta}( X_{t,T} ) ,Y_{t+1,T^{\prime}} )
\end{align}
Here $\mathcal{L}$ is the loss function. In our experiments, we consistently adopt the widely used mean squared error (MSE) function as the loss function.
\section{Methodology}  
In this section, we systematically analyze three primary limitations inherent in the vanilla LIF model when applied to spatio-temporal prediction tasks. To overcome these challenges, we propose a spatio-temporal self-connection circuit model inspired by biological autaptic structures. We elucidate the design principles of the model and provide a comprehensive analysis of its dynamic characteristics.
\subsection{ Limitations of the Vanilla LIF Model in Spatio-temporal Prediction Tasks} 
\textbf{(a): Limited capacity to model long-term dependencies.} 
Spatio-temporal prediction tasks involve capturing temporal features and dependencies to predict future states. The vanilla LIF model employs leakage and discharge processes to regulate historical information, thereby exhibiting certain sequential modeling capabilities. However, its discharge process follows a fixed linear decay, while the leakage process exhibits a fixed exponential decay. To be more specific, according to Eqs.~\eqref{1-1}-\eqref{1-1-4}, the membrane potential of the vanilla LIF model at time-step $t+T$, derived from $\boldsymbol{v}^{l}[t]$, can be obtained after $T$ iterations and is represented as follows:
\begin{align}
\label{1-2}
\boldsymbol{v}^{l}[t+T]&=(\alpha ^l) ^T\boldsymbol{v}^{l}[t]+\sum_{i=1}^T{( \alpha ^l) ^{T-i}( 1-\alpha ^l) \boldsymbol{x}^{l}[t+i]}\nonumber
\\
&-v_{th}\sum_{i=1}^T{(\alpha ^l) ^{T-i}\boldsymbol{s}^{l}[t+i]}
\end{align}
The detailed derivation is provided in the Appendix.
According to Eq.~\eqref{1-2}, we observe an inevitable decay in the historical membrane potential information $\boldsymbol{v}_{t}^{l}$. Specifically,  $\boldsymbol{v}_{t}^{l}$ at time-step $t$ undergoes $T$ steps of exponential decay and $\sum_{i=1}^T{\boldsymbol{s}^{l}}[t+i]$ steps of linear decay after $T$ iterations. As $\alpha ^l$ approaches 0, the decay of membrane potential information becomes more pronounced. Conversely, as $\alpha ^l$ approaches 1, it leads to a weaker input current. Due to the presence of leakage, maintaining membrane potential balance and outputting spike information heavily relies on the current input at the current moment. This monotonic leakage mechanism poses challenges for the vanilla LIF model in capturing long-term dependencies. Moreover, previous studies~\cite{tclif} have highlighted the issue of gradient vanishing when employing BPTT to optimize SNNs with LIF neurons for long-term modeling tasks, further restricting the model's optimization capabilities.

\textbf{(b): Limited capacity to represent spatio-temporal information.} Spatio-temporal prediction necessitates the deep interactions and joint learning of temporal and spatial information, which poses a challenge for LIF model. To see this, we rewrite the  output of the vanilla LIF model at time $t$ as:
\begin{align}
\label{1-x}
\boldsymbol s^l[t] =f( \boldsymbol x^l[t] ,\boldsymbol v^l[t-1] ) =f ( \boldsymbol W^l\boldsymbol s^{l-1}[t] ,\boldsymbol v^l[ t-1 ] ) 
\end{align}
Here, function $f(\cdot)$ denotes the dynamics of the LIF neuron (Eqs.~\eqref{1-1-2}-\eqref{1-1-3}).
For the LIF model, the spiking neurons in the $l$-th layer only receive inputs from the spike outputs of neurons in previous layer, $\boldsymbol{s}^{l-1}[t]$, while the incorporation of historical information is solely based on the neuron's own input at the previous time-step.
Additionally, the network's learnable parameters primarily focus on extracting spatial information, lacking effective parameter training for feature extraction in the temporal domain. This simple connectivity pattern and topology restrict the network's capacity to attain a profound representation of spatial-temporal information.

\textbf{(c): Weak adaptability to different datasets.} Spatio-temporal prediction requires modeling dynamic spatio-temporal data. However, the parameters of the LIF model remain unchanged, whether during training or inference.
Although some variants of the LIF model optimize
parameters like membrane time constants and thresholds in
training~\cite{plif,glif}, these
parameters remain fixed in inference. The distribution and correlation of spatio-temporal data can change over time and space, and fixed parameters are insufficient to capture these variations, resulting in reduced generalization ability in prediction.

\subsection{Spatio-Temporal Circuit Leaky Integrate-and-Fire (STC-LIF) Model}
To overcome the challenges associated with spatio-temporal data learning in vanilla LIF and its variant models, we propose a novel spatio-temporal self-connection circuit model called the STC-LIF model, which can be described as: 
\begin{align}
\label{1-4}
\boldsymbol{x}^{l}[t]&=\boldsymbol{W}^l \boldsymbol{s}^{l-1}[t]
\\
\boldsymbol \beta ^{l}[t]&=\text{Tanh}(\boldsymbol W_{gt}^{l}\boldsymbol{s}^{l}[t-1]) \label{1-4-3}
\\
\boldsymbol \gamma ^{l}[t]&=\text{Tanh}(\boldsymbol W_{gs}^{l}\boldsymbol{s}^{l}[t-1]) 
\\
\boldsymbol{m}^{l}[t]&=\boldsymbol{v}^{l}[t-1]\odot( 1+\boldsymbol \beta ^{l}[t]) +\boldsymbol{x}^{l}[t]\odot( 1+\boldsymbol \gamma ^{l}[t]) 
\\
\boldsymbol{s}^{l}[t]&=H (\boldsymbol{m}^{l}[t]-{v}_{th})\label{1-4-4}
\\
\boldsymbol{v}^{l}[t]&=\boldsymbol{m}^{l}[t]-{v}_{th}\boldsymbol{s}^{l}[t] \label{1-4-1}
\end{align}
Here, $\odot$ denotes the Hadamard product. Compared to the vanilla LIF model presented in Eqs.~\eqref{1-1}-\eqref{1-1-4}, the STC-LIF model retains the integration, spiking, and leakage processes while introducing axon-dendrite and axon-soma circuits inspired by biological autaptic synapses to dynamically regulate the input current and historical
information (Figure~\ref{p1}(c)).
It replaces the fixed leakage mechanism with two dynamic regulatory factors, $\boldsymbol \beta ^{l}[t]$ and $\boldsymbol \gamma ^{l}[t]$. $\boldsymbol W_{gt}^{l}$ and $\boldsymbol W_{gs}^{l}$ represent the synaptic weights of the temporal and the spatial circuits.

We further elaborate on the design principles of STC-LIF. Previous experiments~\cite{au9,au10,au11} have demonstrated that autaptic synapses can regulate the precision and stability of action potential firing by forming delayed feedback that either inhibits or enhances subsequent action potential generation. Therefore, we introduce the regulatory factors, $\boldsymbol \beta ^{l}[t]$ and $\boldsymbol \gamma ^{l}[t]$, to simulate the axon-soma and axon-dendrite circuits found in autaptic structures. These factors are within the range of [-1, 1] and determined by the spike output of the previous time step, $\boldsymbol{s}^{l}[t-1]$. In the axon-dendrite circuit, if $\boldsymbol \gamma ^{l}[t] > 0$, it indicates the dominance of excitatory synapses, resulting in an enhancement of the input current to the pre-synaptic neuron and promoting spiking. Conversely, if $\boldsymbol \gamma ^{l}[t] < 0$, it signifies the dominance of inhibitory synapses, leading to an attenuation of the input current to the pre-synaptic neuron and suppressing spiking. Similarly, in the axon-soma circuit, if $\boldsymbol \beta ^{l}[t] > 0$, it amplifies the membrane potential, thereby preserving significant historical information. Conversely, if $\boldsymbol \beta ^{l}[t] < 0$, it weakens the membrane potential, facilitating the forgetting of irrelevant historical information.

The spatio-temporal circuit enables spiking neurons to regulate the current input and membrane potential at the current time step based on their spike output from the previous time step, aligning with the functionality of autaptic synaptic in biology. However, relying solely on self-spike output for regulation has limitations and inflexibility, potentially constraining the expressive power of SNNs. To address this, we extend the concept of self-connection beyond individual neurons to include connections within the layer of neighboring cells. We use group convolution operation to characterize the influence of the spike output from the neuron itself as well as its adjacent neurons on a spiking neuron. This approach is not only computationally efficient but also more biologically plausible, as it aligns with the sparse connectivity observed in biological networks, where neurons primarily engage in local information exchange rather than global interactions~\cite{guo2017hierarchical}.

\subsection{Dynamic Analysis of STC-LIF Model} 
\textbf{Relate STC-LIF model to LIF and IF models.} The STC-LIF model is a natural extension of the vanilla LIF model. Here we establish the relationship between STC-LIF and the LIF and IF models. According to Eqs.~\eqref{1-4}-\eqref{1-4-1}, when $\boldsymbol \gamma^l[t]$ is less than zero and $\boldsymbol \beta^l[t] +\boldsymbol \gamma^l[t]$ equals $-1$, the STC-LIF model is equivalent to the LIF model. When $\boldsymbol \beta^l[t]=\boldsymbol \gamma^l[t]=0$, the STC-LIF model is equivalent to the IF model.

We further analyze the advantages of the STC-LIF model compared to the LIF and IF models.
The derivation process follows a similar approach to that of Eq.~\eqref{1-2}. According to Eqs.~\eqref{1-4}-\eqref{1-4-1}, the membrane potential of the STC-LIF model at time-step $t+T$, derived from $\boldsymbol{v}^{l}[t]$, can be obtained after $T$ iterations and is represented as follows:
\begin{align}
\label{1-5}
&~~~~\boldsymbol v^l[t+T] 
= \boldsymbol v^l[t]\prod_{j=1}^T{( 1+\boldsymbol\beta ^l[t+j])} \nonumber
\\
&+\sum_{i=1}^T{(1+\boldsymbol\gamma ^l[t+i])\boldsymbol x^l[t+i] \prod_{j=1}^{T-i}{( 1+\boldsymbol\beta ^l[ t+i+j])}} \nonumber
\\
&-\boldsymbol v_{th}\sum_{i=1}^T{\boldsymbol s^l[t+i] \prod_{j=1}^{T-i}{(1+\boldsymbol\beta ^l[t+i+j])}}
\end{align}
The detailed derivation is provided in the Appendix. From Eq.~\eqref{1-5}, the STC-LIF model has the following advantages:

\textbf{(a): STC-LIF model can capture long-term dependency.} From Eq.~\eqref{1-5}, it is evident that the historical membrane potential information is determined by the dynamic modulation factors $\boldsymbol \beta^l[t]$ and $\boldsymbol \gamma^l[t]$. For the membrane potential $\boldsymbol v^l[t]$ at time-step $t$, we can effectively transmit it to the membrane potential $\boldsymbol v^l[t+T]$ at $T$ steps later by employing appropriate dynamic modulation. The factor $\boldsymbol \gamma^l[t]$ also plays a vital role in regulating the current input at the current time-step, thereby ensuring a relative balance between the input and the historical membrane potential. Moreover, the spatio-temporal circuit plays a critical role in preserving the propagation of gradient errors, which is fundamental for optimizing SNNs using the BPTT method. The temporal flow of gradient information in the STC-LIF model can be represented by the following recursive form:
\begin{align}
&~~\small{\frac{\partial\boldsymbol v^l[t+T]}{\partial \boldsymbol v^l[ t]}}=\prod_{i=1}^T{\small{\frac{\partial \boldsymbol v^l[t+i]}{\partial \boldsymbol m^l[ t+i]}\frac{\partial \boldsymbol m^l[t+i]}{\partial \boldsymbol v^l[t+i-1]}}}
\\
&=\prod_{i=1}^T{(1-v_{th}H^\prime(\boldsymbol m^l[ t+i] -v_{th})) (1+\boldsymbol\beta ^l[ t+i])} \nonumber
\end{align}
The derivative $H^\prime(\cdot)$ is often approximated by the derivative of a continuous surrogate function.
As $T$ increases, it is possible that $\prod_{i=1}^T{(1-v_{th}H^\prime(\boldsymbol m^l[ t+i] -v_{th}))}\rightarrow 0 $ or $\prod_{i=1}^T{(1-v_{th}H^\prime(\boldsymbol m^l[ t+i] -v_{th}))}\rightarrow \infty $, which may lead to gradient vanishing and exploding issues.
However, by appropriately assigning the values of $\boldsymbol\beta ^l[t]$, we can effectively alleviate the problems, thereby preserving the propagation of gradient flow information.

\textbf{(b): STC-LIF model enhances spatio-temporal representation capability.} The spike output of the STC-LIF model for layer $l$ can be represented as follows:
\begin{align}
\label{1-x}
\boldsymbol s^l[t] =f ( \boldsymbol W^l\boldsymbol s^{l-1}[t],\boldsymbol W_{gs,gt}^{l}\boldsymbol s^l[t-1],\boldsymbol v^l[ t-1 ] ) 
\end{align}
Here, the function $f(\cdot)$ represents the dynamics of the STC-LIF neuron, as depicted in Eqs.~\eqref{1-4-3}-\eqref{1-4-4}. The spike output of the STC-LIF model is influenced by both asynchronous signals within the layer and synchronous signals between layers. The spatio-temporal circuit forms new synaptic connections, thereby enhancing the interaction of spatio-temporal information. The learnable synaptic weights $\boldsymbol W_{gs}^{l}$ and $\boldsymbol W_{gt}^{l}$ strengthen the capability to extract spatial and temporal features, consequently advancing the representational capacity of spiking neurons.

\begin{figure}[t]
\vskip 0.2in
\begin{center}
\centerline{\includegraphics[width=\columnwidth]{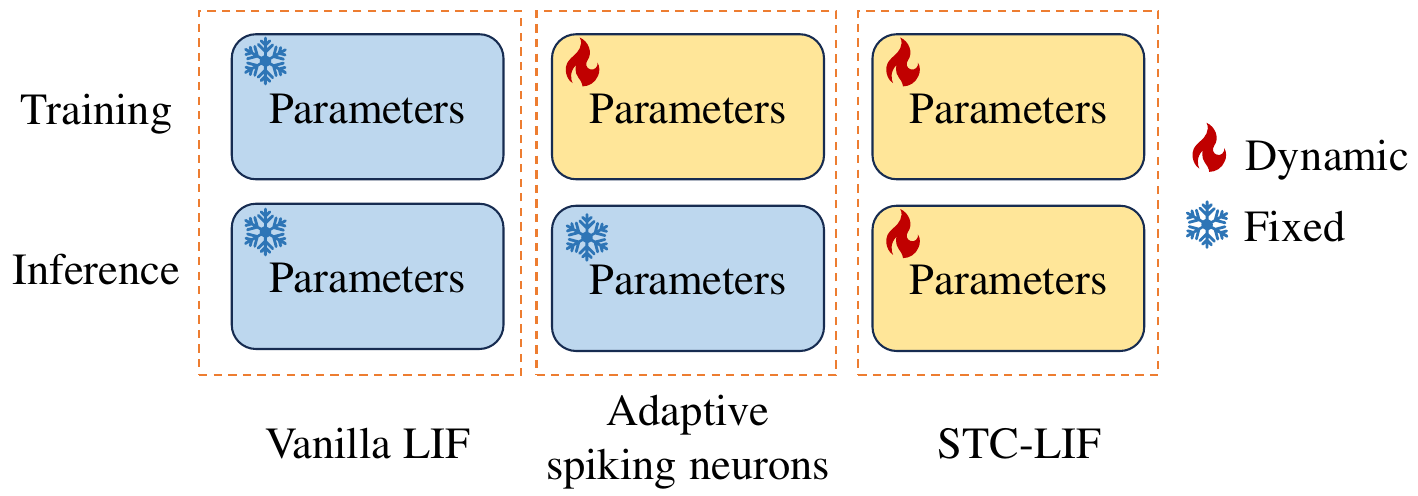}}
\caption{Comparison of parameter changes of different models.}

\label{p2}
\end{center}
\vskip -0.2in
\end{figure}

\textbf{(c): STC-LIF model has dynamic parameters.} 
As shown in \cref{p2}, the parameters of the STC-LIF model undergo dynamic changes during both training and inference. The current parameters of the neuron are influenced by the feedback modulation from its output in the previous time step. In essence, the parameters of the STC-LIF model are adaptive and evolve with the input sequence. This enables the generation of diverse dynamics parameters, catering to different input sequences and thereby enhancing the network's adaptability to various tasks and scenarios. Consequently, the model exhibits improved self-adaptiveness.

\section{Experiments}
In this section, we conducted experiments on multiple spatio-temporal prediction benchmark tasks to validate the effectiveness of our proposed method. These tasks include the Moving MNIST~\cite{mmnist} dataset, the TaxiBJ~\cite{taxibj} traffic flow dataset, and the KTH~\cite{kth} human action dataset. We compare our method with the Vanilla LIF model, and adaptive spiking neural models, including PLIF~\cite{plif}, GLIF~\cite{glif}, TC-LIF~\cite{tclif} and LM-H~\cite{lmh}.

\textbf{Datasets.} 
Moving MNIST is a video sequence dataset generated from the MNIST dataset, each frame of which contains two randomly moving digits. The objective of this dataset is to input the first 10 frames and predict the positions and shapes of the digits in the subsequent 10 frames. TaxiBJ is a dataset that contains GPS data from taxis in Beijing. The objective is to predict future urban population flow based on historical data. KTH is a dataset designed for human action recognition. The goal of this dataset is to recognize human action categories based on video sequences. The statistical summaries of these datasets are provided in Table.~\ref{table-11}.

\begin{table}[t]
\caption{The statistics of datasets. The training/testing set has
$N_{train}$/$N_{test}$ samples, composed of $T$ and $T^{\prime}$ images.} 
\centering
\vskip 0.1in
\resizebox{\columnwidth}{!}{
\begin{tabular}{cccccc}
   \toprule
 &$N_{train}$&$N_{test}$&$\left( C,H,W \right)$&$T$&$T^{\prime}$\\
   \midrule
    Moving MNIST&10000&10000&(1,64,64)&10&10\\
    TaxiBJ&19627&1334&(2,32,32)&4&4\\
    KTH&5200&3167&(1,128,128)&10&20 or 40\\
   \bottomrule
\end{tabular}}
\vskip -0.1in
\label{table-11}
\end{table}

\textbf{Evaluation.} We evaluated the prediction performance using commonly used metrics in spatio-temporal prediction tasks, namely mean squared error (MSE), mean absolute error (MAE), structural similarity index measure (SSIM), and peak signal-to-noise ratio (PSNR).

\begin{figure*}[t]
\vskip 0.2in
\begin{center}
\centerline{\includegraphics[width=0.97\linewidth]{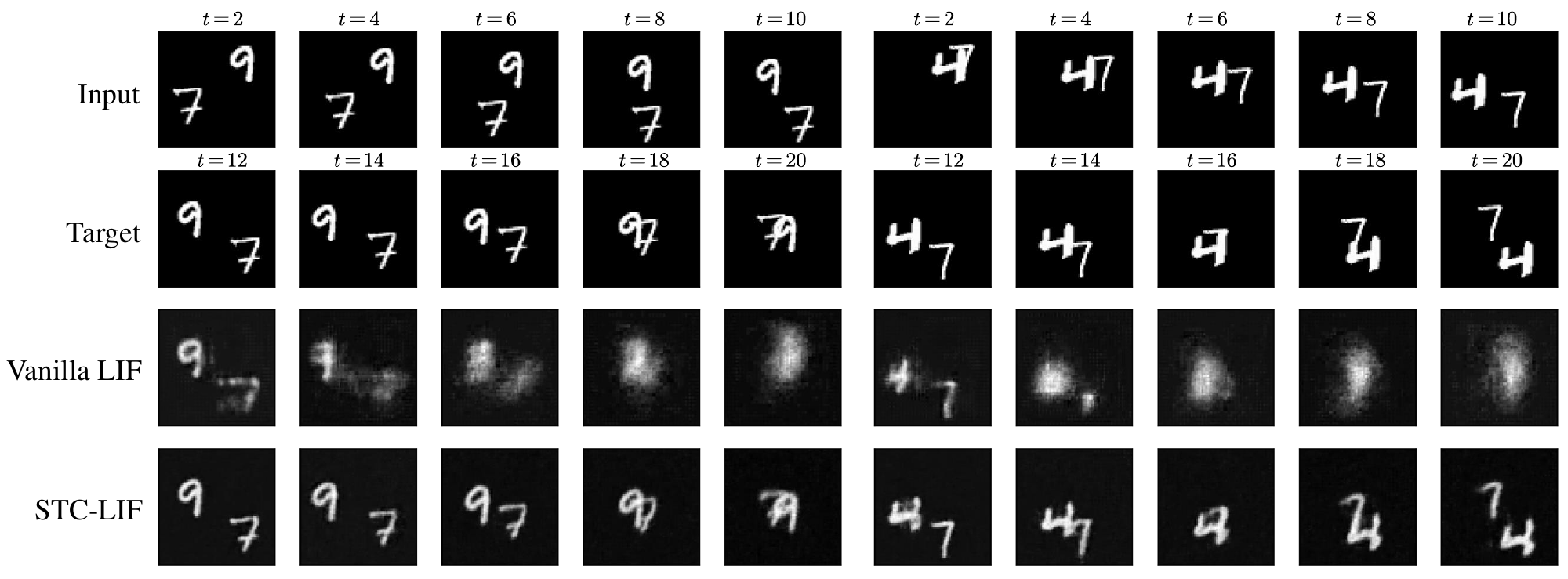}}
\caption{Qualitative visualization of the prediction results of vanilla LIF and STC-LIF models on the Moving MNIST dataset.}
\label{p3s1}
\end{center}
\vskip -0.2in
\end{figure*}

\textbf{Detailed Setup.}Our network architecture consisted of four layers, with the first three layers composed of spiking neurons and channel dimensions set to $[256, 256, 256]$. The feature map resolution remained constant throughout networks. The final layer was a convolutional layer designed to restore the predicted image dimensions. 
In order to address covariate shift~\cite{gn} and enhance training stability, each convolutional layer, except for the last one, was followed by a group normalization layer.
For more detailed information regarding the experiments, please refer to the Appendix.

\subsection{Evaluation on the Moving MNIST Dataset} 

\begin{table}[t]
\caption{Quantitative results on the Moving MNIST dataset.}
\centering
\vskip 0.1in
\resizebox{\columnwidth}{!}{
\begin{tabular}{cccc}
   \toprule

    Method&MSE↓&MAE↓&SSIM↑\\
   \midrule 
    Vanilla LIF&102.8&246.2&0.640\\    
    PLIF~\cite{plif}&99.2&233.3&0.691\\    
    GLIF~\cite{glif}&79.7&225.3&0.685\\ 
    TCLIF~\cite{tclif}&105.1&257.9&0.634\\    
    LM-H~\cite{lmh}&93.2&244.5&0.661\\    
    \textbf{STC-LIF}&\textbf{47.0}&\textbf{136.4}&\textbf{0.863}\\

   \bottomrule
\end{tabular}
}
\vskip -0.1in
\label{table2}
\end{table}

Moving MNIST is a benchmark dataset for spatio-temporal prediction.It poses challenges due to the random motion trajectories of the digits, which encompass complex scenarios like collisions, occlusions, and boundary rebounds. Therefore, predicting future frames necessitates modeling and inferring information such as digit shape, speed, direction, and position. Additionally, the dataset's low resolution introduces blurriness, further complicating the prediction.

Quantitative results for Moving MNIST are presented in Table~\ref{table2}. It is evident that the STC-LIF model outperforms other spiking neuron models in terms of MSE, MAE, and SSIM metrics. For instance, our model reduces the MSE by 55.8 compared to the vanilla LIF model and by 32.7 compared to the best-performing PLIF model among other spiking neuron models.
Qualitative prediction results are depicted in Figure~\ref{p3s1}. The images generated by the vanilla LIF model are notably blurry, particularly in later time steps. Conversely, the STC-LIF model produces higher-quality images, clearly depicting the trajectories and contours of each digit. Therefore, it can be concluded that the STC-LIF model significantly enhances the extraction of spatio-temporal features and the modeling of long-term dependencies. Additional visualization results can be found in the Appendix.

\subsection{Evaluation on the TaxiBJ Dataset} 

\begin{table}[t]
\caption{Quantitative results on the TaxiBJ dataset.}
\centering
\vskip 0.1in
\resizebox{\columnwidth}{!}{
\begin{tabular}{cccc}
   \toprule
    Method&MSE$(\times100)$↓&MAE↓&SSIM↑\\
   \midrule
 
    Vanilla LIF&70.4&22.1&0.966\\    
    PLIF~\cite{plif}&70.8&21.9&0.967\\    
    GLIF~\cite{glif}&62.9&20.6&0.971\\
    TCLIF~\cite{tclif}&59.2&19.3&0.973\\    
    LM-H~\cite{lmh}&63.3&20.1&0.971\\    
    \textbf{STC-LIF}&\textbf{56.4}&\textbf{19.5}&\textbf{0.973}\\

   \bottomrule
\end{tabular}
}
\label{table3}
\vskip -0.1in
\end{table}

TaxiBJ is a real and representative urban traffic flow dataset that holds significant value in urban planning and traffic management. Traffic patterns are known to be complex and highly dynamic, with strong dependencies between adjacent timestamps. In the TaxiBJ dataset, each data sequence consists of 8 consecutive frames, and the task involves predicting the remaining 4 frames given the initial 4 frames as input. We evaluated the performance of our model on this dataset, and the quantitative results are reported in Table~\ref{table3}. Additionally, we assessed the MSE differences $(\times100)$ for each frame, as presented in Table~\ref{table4}. It is evident that among all the spiking neuron models, our proposed STC-LIF model demonstrated the best predictive performance, exhibiting the smallest prediction errors.

\begin{table}[t]
\caption{ Per-frame MSE$(\times100)$ on the TaxiBJ dataset. }
\centering
\vskip 0.1in
\resizebox{\columnwidth}{!}{
\begin{tabular}{ccccc}
   \toprule
    Method&Frame1&Frame2&Frame3&Frame4\\
   \midrule
    Vanilla LIF&66.7&68.1&68.5 &78.5 \\    
    PLIF~\cite{plif}&66.8&66.9&71.5&78.0\\    
    GLIF~\cite{glif}&55.3&60.5&65.5 &78.4\\
    TCLIF~\cite{tclif}&49.0&56.8&62.2&69.2\\    
    LM-H~\cite{lmh}&59.7&62.5&64.4&67.6 \\    
    \textbf{STC-LIF}&\textbf{48.1}&\textbf{54.8}&\textbf{58.9}&\textbf{63.0}\\

   \bottomrule
\end{tabular}
}
\label{table4}
\vskip -0.1in
\end{table}

\begin{table}[t]
\caption{Quantitative results on the KTH dataset.}
\centering
\vskip 0.1in
\resizebox{\columnwidth}{!}{
\begin{tabular}{ccccc}
   \toprule
&\multicolumn{2}{c}{KTH$(10\rightarrow 20)$}& \multicolumn{2}{c}{KTH$( 10\rightarrow 40)$} \\
   \midrule[0pt]
    Method&SSIM↑&PSNR↑&SSIM↑&PSNR↑\\
   \midrule
     
    Vanilla LIF&0.767&21.62&0.754&20.89\\    
    PLIF~\cite{plif}&0.796&21.79&0.785&21.17\\    
    GLIF~\cite{glif}&0.805&22.41&0.795&21.90\\
    TCLIF~\cite{tclif}&0.786&22.08&0.746&20.0\\    
    LM-H~\cite{lmh}&0.710&17.13&0.708&16.91\\    
    \textbf{STC-LIF}&\textbf{0.822}&\textbf{23.14}&\textbf{0.815}&\textbf{22.63}\\
 
   \bottomrule
\end{tabular}
}
\vskip -0.1in
\label{table5}
\end{table}

\subsection{Evaluation on the KTH Dataset} 

\begin{table*}[t]
\caption{Comparison of quantitative results of the adaptive spiking neural models and the enhanced models.}
\centering
\vskip 0.1in
\resizebox{\linewidth}{!}{
\begin{tabular}{ccccccccccc}
   \toprule
    &\multicolumn{3}{c}{Moving MNIST}& \multicolumn{3}{c}{TaxiBJ}&\multicolumn{2}{c}{KTH$(10\rightarrow 20)$}& \multicolumn{2}{c}{KTH$(10\rightarrow 40)$} \\
   \midrule[0pt]
Method&MSE↓&MAE↓&SSIM↑&MSE×100↓&MAE↓&SSIM↑&SSIM↑&PSNR↑&SSIM↑&PSNR↑\\
   \midrule
PLIF~\cite{plif}&99.2&233.3&0.691&70.8&21.9&0.967&0.796&21.79&0.785&21.17\\  

\textbf{STC-PLIF}&\textbf{48.4}&\textbf{139.4}&\textbf{0.860}&\textbf{58.6}&\textbf{20.0}&\textbf{0.972}&\textbf{0.819}&\textbf{23.02}&\textbf{0.806}&\textbf{22.32}
\\
LM-H~\cite{lmh}&93.2&244.5&0.661&63.3&20.1&0.971&0.710&17.13&0.708&16.91\\    
  
\textbf{STC-LM-H}&\textbf{45.9}&\textbf{125.8}&\textbf{0.881}&\textbf{54.6}&\textbf{19.3}&\textbf{0.975}&\textbf{0.827}&\textbf{23.16}&\textbf{0.818}&\textbf{22.61}
\\

   \bottomrule
\end{tabular}
}
\label{table6}
\vskip -0.1in
\end{table*}

Recursive models offer the advantage of flexible output length expansion. We evaluate the dynamic generalization of our model for predicting variable frame numbers on the KTH dataset.
The task involved predicting the next 20 frames or 40 frames given a 10-frame input. During training, we made predictions for 20 frames, but during evaluation, we extended it to predict 40 frames, distinguishing it from the previous two datasets. We used SSIM and PSNR as evaluation metrics, where higher values indicate higher quality predictions.
The results are presented in Table~\ref{table5}. Notably, our model exhibited superior performance compared to other spiking neuron models in both scenarios: predicting 20 frames and extending the prediction to 40 frames.

\subsection{Generalize to Other Spiking Neuron Models}


The concept of the spatio-temporal self-connection circuit is highly adaptable and scalable, making it compatible with various spiking neural models~\cite{yu2018emergent,yu2018winner}. Here we extend its application to the PLIF and LM-H models, denoting the enhanced versions as STC-PLIF and STC-LM-H, respectively. The detailed derivations can be found in the Appendix.
Table~\ref{table6} presents the results, clearly indicating that both STC-PLIF and STC-LM-H outperform their respective base models, PLIF and LM-H, by a significant margin. Notably, the improvement is particularly pronounced for the LM-H model, which has more complex dynamics.
These results demonstrate the generality of our model and its effectiveness in enhancing spiking neurons' ability to extract spatio-temporal features and model long-term dependencies.

\begin{figure}[t]
\vskip 0.1in
\begin{center}
\subfigure[Vanilla LIF]{
\includegraphics[width=0.48\columnwidth]{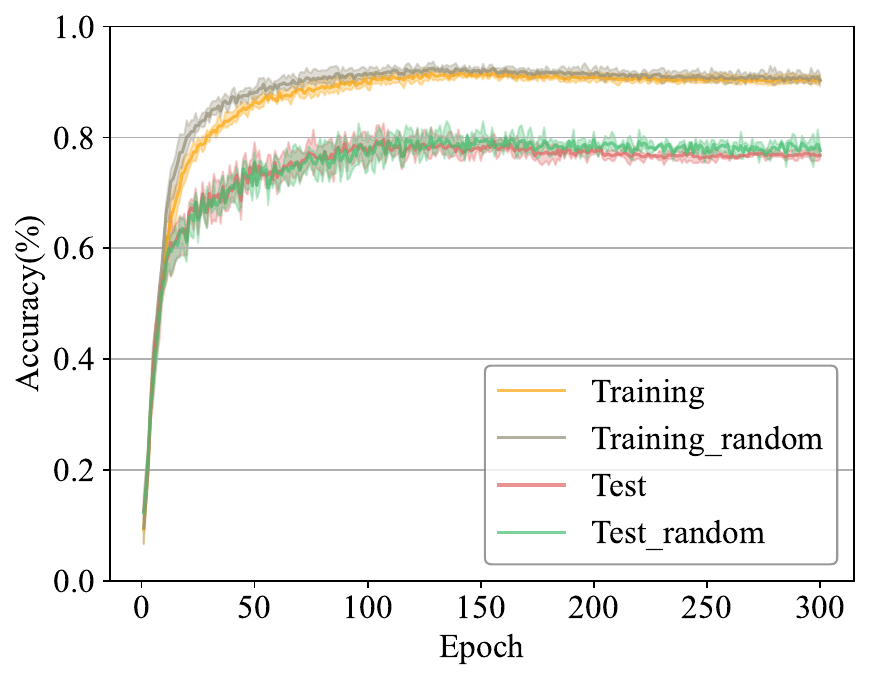}}
\subfigure[STC-LIF]{
\includegraphics[width=0.48\columnwidth]{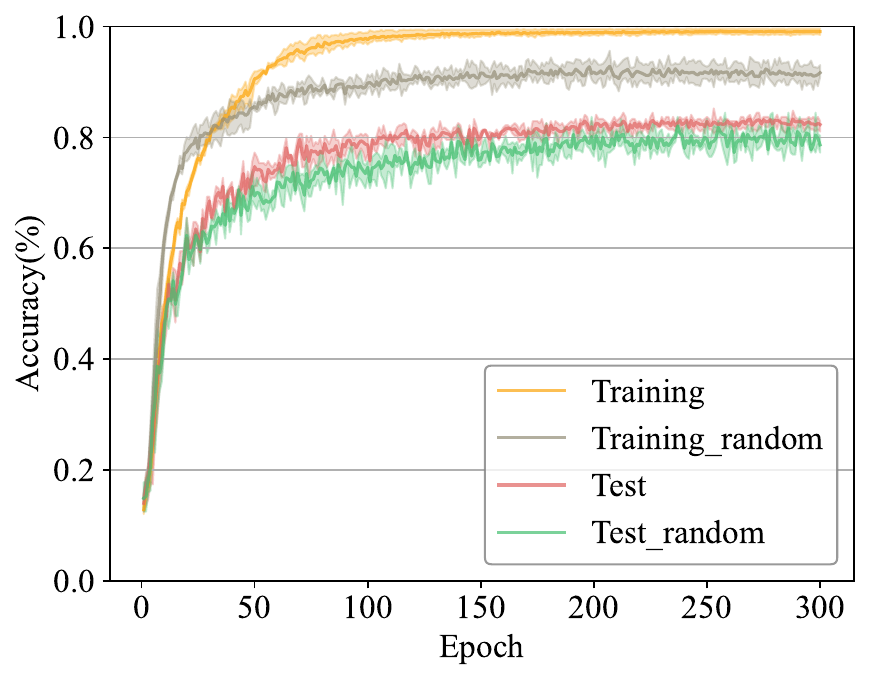}}
\caption{Comparison of vanilla LIF (a) and STC-LIF (b) when inputting sequential sequences or random sequences from the DVS128 Gesture dataset. The experiments were repeated three times, and the solid curve represents the average value.}

\label{p5}
\end{center}
\vskip -0.2in
\end{figure}

\subsection{Rethinking Neuromorphic Data Recognition} 
The neuromorphic dataset contains asynchronous, event-driven, and sparse spatio-temporal information, making it naturally suitable for processing with spiking neural networks. However, can SNNs truly learn the rich spatio-temporal information in neuromorphic data? Here we consider the commonly used DVS128 Gesture~\cite{dvs128} dataset to evaluate the performance of SNNs. The dataset provides information such as timestamps, spatial coordinates, and polarity, encompassing both spatial (gesture positions) and temporal (gesture order) aspects. During training, we use the normal gesture sequence as input, but during prediction, we deliberately disrupt the gesture order to discern the information SNNs rely on for classification. As shown in Figure~\ref{p5}, the vanilla LIF model achieves comparable performance on randomly shuffled input and sequential input, implying that it primarily learns spatial positional relationships. In contrast, the STC-LIF model exhibits a significant decrease in accuracy for randomly shuffled input across both training and testing sets. This indicates that the STC-LIF model is capable of simultaneously learning both temporal and spatial feature in neuromorphic data.

\begin{table}[t]
\caption{Ablation study of our proposed method.}
\centering
\vskip 0.1in
\resizebox{\columnwidth}{!}{
\begin{tabular}{ccccc}
   \toprule

    Method&MSE↓&MAE↓&Parameters(M)&Flops(G)\\
   \midrule
    Vanilla LIF (Baseline)&102.8&246.2&3.305&64.346\\    
    STC-LIF w/o TC&62.5&168.1&3.614&70.398\\    
    STC-LIF w/o SC&58.3&163.8&3.614&70.398\\    
    STC-LIF&47.0&136.4&3.922&76.449\\
   \bottomrule
\end{tabular}
}
\label{table7}
\vskip -0.1in
\end{table}

\subsection{Computational Cost and Ablation Study}

We perform ablative experiments on the STC-LIF model and present the results in Table~\ref{table7}.The findings indicate that both the temporal circuit (TC) and spatial circuit (SC) contribute to the model's performance, with the temporal circuit playing a more prominent role. In terms of computational cost, the spatio-temporal circuit exhibits a modest increase in parameter count and computational complexity per module, thanks to the utilization of grouped convolutions, with respective increments of only 9.3\% and 9.4\%. Considering the noticeable improvement in performance, the increase in computational cost is clearly acceptable.

\section{Conclusion}
This paper analyzes the limitations of the vanilla LIF model in spatio-temporal prediction tasks. Inspired by biological autaptic synapses, we propose the STC-LIF model, which incorporates a spatio-temporal self-connection circuit to enhance the extraction of spatio-temporal features and model spatio-temporal dependencies. Our method achieves state-of-the-art performance among all spiking neuronal models, showcasing the immense potential of temporal processing capabilities in SNNs. 

\section*{Acknowledgements}
This work was supported by the National Natural Science Foundation of China (62176003, 62088102) and the Beijing Nova Program (20230484362).

\section*{Impact Statement}
This paper presents work whose goal is to advance the field of Machine Learning. There are many potential societal consequences of our work, none which we feel must be specifically highlighted here.

\nocite{langley00}

\bibliography{example_paper}
\bibliographystyle{icml2024}

\newpage
\appendix
\onecolumn
\section{Experimental Details}
\textbf{Datasets.} Moving MNIST~\cite{mmnist} serves as a benchmark dataset for spatio-temporal prediction. It is a video sequence dataset generated from the MNIST dataset. It comprises 10,000 sequences, each consisting of 20 frames. Each frame is a grayscale image with dimensions 64x64 and contains two randomly moving digits. 
The objective of this dataset is to input the first 10 frames and predict the positions and shapes of the digits in the subsequent 10 frames.
TaxiBJ~\cite{taxibj} is a dataset that contains GPS data from taxis in Beijing, including both inflow and outflow channels. 
The objective is to predict future urban population flow based on historical data.
KTH~\cite{kth} is a dataset designed for human action recognition. It includes six different actions (walking, jogging, running, boxing, waving, and clapping) performed by 25 individuals in four distinct scenarios (outdoor, outdoor with scale variation, outdoor with clothing variation, and indoor). 
The goal of this dataset is to recognize human action categories based on video sequences. 
We used individuals 1-16 for training and individuals 17-25 for prediction.

\textbf{Evaluation.} We evaluated the prediction performance using commonly used metrics in spatio-temporal prediction tasks, namely mean squared error (MSE), mean absolute error (MAE), structural similarity index measure (SSIM), and peak signal-to-noise ratio (PSNR).
MSE and MAE are image-based metrics that measure the similarity of content in the spatial domain. They calculate the average of squared or absolute pixel differences between the ground truth and predicted images, respectively.
SSIM is a feature-based metric that evaluates category/texture similarity. It mimics the theory of structural similarity in the human visual system (HVS) and demonstrates sensitivity to local structural variations in images. SSIM quantifies image attributes such as luminance, contrast, and structure by estimating the mean for luminance, variance for contrast, and covariance for structural similarity. SSIM values range from 0 to 1, with higher values indicating greater image similarity. One advantage of SSIM is its better alignment with human subjective evaluation.
PSNR is a widely used measure for evaluating image quality. It represents the reference value of image quality between the maximum signal and background noise. PSNR is expressed in decibels (dB), with higher values indicating lesser image distortion.

\begin{figure}[htbp]
\vskip 0.2in
\begin{center}
\centerline{\includegraphics[width=0.5\columnwidth]{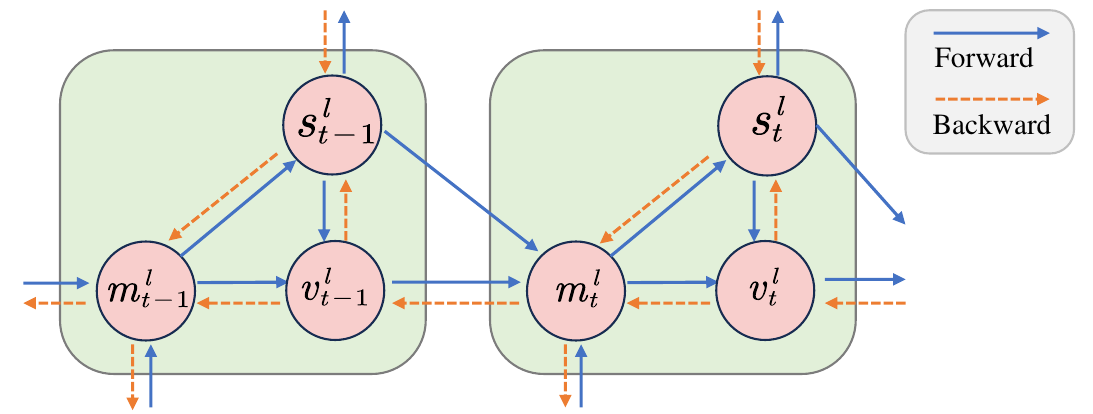}}
\caption{The unfolded computation graph of STC-LIF model.}
\label{p4}
\end{center}
\vskip -0.2in
\end{figure}

\textbf{Decoupling of spike output and self-regulation.}
In the STC-LIF model, the spiking signals play a dual role in information output and self-regulation. Therefore, it is necessary to decouple these two functions to strike a balance between the output and regulation of spiking neurons. During backpropagation, we truncate the gradient information of the spatio-temporal circuit (Figure~\ref{p4}). This ensures that the additional learnable weight parameters, $\boldsymbol W_{gt}^{l}$ and $\boldsymbol W_{gs}^{l}$, only affect the optimization of the dynamic modulation factors, $\boldsymbol\beta ^l[ t]$ and $\boldsymbol\gamma ^l[t]$, without impacting the output of the spiking signals.

\textbf{Implementation Details.} The experimental settings of different datasets are as shown in Table~\ref{table8}. For the Moving MNIST and KTH datasets, we set the patchsize to $2\times2$ and $4\times4$, which means that each $64\times64$ frame and $128\times128$ frame are resized into $4\times32\times32$ and $16\times32\times32$ tensors for the representation, following~\cite{mmnist}. The kernel size of all convolutional layers is set to $5\times5$, stride to 1 and padding to 2. For experiments on the DVS128 Gesture dataset, we used a 3Conv-1FC classification architecture. The Conv layer consisted of a convolutional layer, a spiking neuron layer, and a Maxpool layer, with dimensions of [64,64,64]. The FC layer contained an AvgPool and a linear fully connected layer. Each Maxpool layer and Avgpool layer performed a downsampling of the feature map by a factor of 2.The Adam~\cite{adam} optimizer was employed for all experiments, and we utilized a combination of warm-up~\cite{warmup} and cosine learning rate scheduling~\cite{cos}. The other hyperparameters are shown in Table~\ref{table9}. The hyperparameters of the PLIF~\cite{plif}, GLIF~\cite{glif}, TC-LIF~\cite{tclif} and LM-H~\cite{lmh} models follow the settings in their papers or public codes. Our codes are available \href{https://github.com/wangtianyi1874/stclif}{https:\textbackslash\textbackslash github.com\textbackslash wangtianyi1874\textbackslash stclif}. The implementation is based on the OpenSTL~\cite{tan2023openstl} and SpikingJelly~\cite{fang2023spikingjelly} frameworks, utilizing a single NVIDIA-3090 GPU.

\begin{table*}[htbp]
\caption{Experimental settings of different datasets.}
\centering
\vskip 0.1in
\resizebox{\linewidth}{!}{
\begin{tabular}{ccccccc}
   \toprule
   Dataset&Architecture&Resize&Optimizer(weight decay)&Batchsize&Epochs&Gradient Detach\\ 
   \midrule
    Moving MNIST&4Conv&(4,32,32)&adam(0)&16&500&N\\        
    TaxiBJ&4Conv&(2,32,32)& adam(0) &16&200&Y \\        
    KTH&4Conv& (16,32,32)& adam(0) &16&200&Y \\        
    DVS128 Gesture&3Conv-FC&(2,128,128)& sgd(0.05) &8&300&Y \\
   \bottomrule
\end{tabular}
}
\label{table8}
\vskip -0.1in
\end{table*}

\begin{table}[htbp]
\caption{Hyperparameter settings for ours experimental implementation.}
\centering
\vskip 0.1in
\resizebox{0.5\columnwidth}{!}{
\begin{tabular}{ccc}
   \toprule
   Parameters&Descriptions&Value\\ 
   \midrule
    $\alpha ^l$&membrane constant&0.5\\        
    $v_{th}$&firing threshold&1.0\\        
    -&initial learning rate&1e-3\\        
    -&final learning rate&1e-5\\  
    -&learning rate schedule&cosine decay\\
    -&warmup epochs&10\\             
    $\gamma$& sgd optimizer momentum&0.9\\        
    $\beta_{1},\beta_{2}$& adam optimizer momentum&0.9,0.999\\
    \multirow{2}{*}{$k$}& number of groups for group&\multirow{2}{*}{16} \\ 
    &convolution and group normalization&\\ 
    \multirow{2}{*}{$n$}& number of frames for dvs128&\multirow{2}{*}{20}\\
    &gesture dataset sliced along time axis&\\

   \bottomrule
\end{tabular}
}
\label{table9}
\vskip -0.1in
\end{table}

\section{Proof of Theorem}

\textbf{Proof of Eq.~\eqref{1-2}.} According to Eqs.~\eqref{1-1}-\eqref{1-1-4},We derived the relationship between the membrane potential $\boldsymbol{v}^{l}[t]$ at time $t$ and the membrane potential $\boldsymbol{v}^{l}[t+T]$ after $T$ iterations of the vanilla LIF model. The iterative derivation process is represented as follows:
\begin{align}
\label{7.1}
\boldsymbol{v}^{l}[t+T]&=\alpha ^l\boldsymbol v^{l}[t+T-1]+(1-\alpha ^l) \boldsymbol x^l[t+T]-v_{th}\boldsymbol s^l[t+T] \nonumber
\\
&=( \alpha ^l ) ^2\boldsymbol v^l[t+T-2] +\alpha ^l( 1-\alpha ^l ) \boldsymbol x^l[t+T-1] -v_{th}\alpha ^l\boldsymbol s^l[t+T-1] +(1-\alpha ^l)\boldsymbol x^l[ t+T ] -v_{th}\boldsymbol s^l[ t+T] \nonumber
\\
&=(\alpha ^l) ^T\boldsymbol{v}^{l}[t]+\sum_{i=1}^T{( \alpha ^l) ^{T-i}( 1-\alpha ^l) \boldsymbol{x}^{l}[t+i]}-v_{th}\sum_{i=1}^T{(\alpha ^l) ^{T-i}\boldsymbol{s}^{l}[t+i]}
\end{align}


\textbf{Proof of Eq.~\eqref{1-5}.} Similarly,for the STC-LIF model, according to Eqs.~\eqref{1-4}-\eqref{1-4-1}, we derived the relationship between the membrane potential $\boldsymbol{v}^{l}[t]$ at time $t$ and the membrane potential $\boldsymbol{v}^{l}[t+T]$ after $T$ iterations. The iterative derivation process is represented as follows:
\begin{align}
\label{7.2}
\boldsymbol v^l[t+T]&=(1+\boldsymbol\beta ^l[ t+T])\boldsymbol v^l[t+T-1]+(1+\boldsymbol\gamma ^l[t+T])\boldsymbol x^l[t+T] -v_{th}\boldsymbol s^l[t+T]\nonumber
\\
&=(1+\boldsymbol\beta ^l[t+T])(1+\boldsymbol\beta ^l[t+T-1]) \boldsymbol v^l[t+T-2]+(1+\boldsymbol\beta ^l[t+T])(1+\boldsymbol\gamma ^l[t+T-1]) \boldsymbol x^l[t+T-1]\nonumber
\\
&-\boldsymbol v_{th}(1+\boldsymbol\beta ^l[t+T])\boldsymbol s^l[t+T-1]+(1+\boldsymbol\gamma ^l[t+T])\boldsymbol x^l[t+T] -v_{th}\boldsymbol s^l[t+T]\nonumber
\\
&=\boldsymbol v^l[t] \prod_{j=1}^T{( 1+\boldsymbol\beta ^l[t+j])}+\sum_{i=1}^T{(1+\boldsymbol\gamma ^l[t+i])\boldsymbol x^l[t+i]}\prod_{j=1}^{T-i}{( 1+\boldsymbol\beta ^l[ t+i+j])}\nonumber
\\
&-\boldsymbol v_{th}\sum_{i=1}^T{\boldsymbol s^l[t+i]\prod_{j=1}^{T-i}{(1+\boldsymbol\beta ^l[t+i+j])}}
\end{align}

\section{Enhanced Versions of PLIF and LM-H Models}

\textbf{The STC-PLIF Model.} We extended the spatio-temporal circuit to the PLIF model and the LM-H model. The enhanced version for the PLIF model, STC-PLF model, is represented as follows:

\begin{align}
\boldsymbol{x}^{l}[t]&=\boldsymbol{W}^l \boldsymbol{s}^{l-1}[t]
\\
\textcolor{red}{\boldsymbol \beta ^{l}[t]}&\textcolor{red}{=\text{Tanh}(\boldsymbol W_{gt}^{l}\boldsymbol{s}^{l}[t-1])}
\\
\textcolor{red}{\boldsymbol \gamma ^{l}[t]}&\textcolor{red}{=\text{Tanh}(\boldsymbol W_{gs}^{l}\boldsymbol{s}^{l}[t-1])} 
\\
\boldsymbol{m}^{l}[t]&=\alpha ^l\boldsymbol{v}^{l}[t-1]\textcolor{red}{\odot( 1+\boldsymbol \beta ^{l}[t])} +(1-\alpha ^l)\boldsymbol{x}^{l}[t]\textcolor{red}{\odot( 1+\boldsymbol \gamma ^{l}[t])} 
\\
\boldsymbol{s}^{l}[t]&=H (\boldsymbol{m}^{l}[t]-{v}_{th})
\\
\boldsymbol{v}^{l}[t]&=\boldsymbol{m}^{l}[t]-{v}_{th}\boldsymbol{s}^{l}[t] 
\end{align}

Here the black part is the original formula of the PLIF model, and the red part is the improvement for the PLIF model. It is worth noting that the membrane constant $\alpha ^l$ is different from the vanilla LIF model in that it is a learnable parameter, and its range is [0,1]. The PLIF model enhances the adaptability of spiking neurons by learnable membrane constant. Our improvement for the PLIF model is similar to the vanilla LIF model, which adds the regulatory factors $\boldsymbol \beta ^{l}[t]$ and $\boldsymbol \gamma ^{l}[t]$ on the basis of the original dynamics of the PLIF model.

\textbf{The STC-LM-H Model.} The enhanced version for the LM-H model, STC-LM-H model, is represented as follows:

\begin{align}
\boldsymbol{x}^{l}[t]&=\boldsymbol{W}^l \boldsymbol{s}^{l-1}[t]
\\
\textcolor{red}{\boldsymbol \beta ^{l}[t]}&\textcolor{red}{=\text{Tanh}(\boldsymbol W_{gt}^{l}\boldsymbol{s}^{l}[t-1])}
\\
\textcolor{red}{\boldsymbol \gamma ^{l}[t]}&\textcolor{red}{=\text{Tanh}(\boldsymbol W_{gs}^{l}\boldsymbol{s}^{l}[t-1])} 
\\
\boldsymbol{v}_{D}^{l}[t]&=\mu _{D}^{l}\boldsymbol{v}_{D}^{l}[t-1]+ \mu _{S}^{l}\boldsymbol{v}_{S}^{l}[t-1]+\boldsymbol{x}^{l}[t]\textcolor{red}{\odot( 1+\boldsymbol \gamma ^{l}[t])}
\\
\boldsymbol{m}^{l}[t]&=\lambda _{S}^{l}\boldsymbol{v}_{S}^{l}[t-1]\textcolor{red}{\odot( 1+\boldsymbol \beta ^{l}[t])}+ \lambda _{D}^{l}\boldsymbol{v}_{D}^{l}[t]
\\
\boldsymbol{s}^{l}[t]&=H (\boldsymbol{m}^{l}[t]-{v}_{th})
\\
\boldsymbol{v}_{S}^{l}[t]&=\boldsymbol{m}^{l}[t]-{v}_{th}\boldsymbol{s}^{l}[t] 
\end{align}

Similarly, the black part is the original formula of the LM-H model, and the red part is the improvement for the LM-H model. The LM-H model has more complex neural dynamics than the vanilla LIF model and the PLIF model. It is a two-compartment model, which contains two membrane potentials $\boldsymbol{v}_{D}^{l}[t]$ and $\boldsymbol{v}_{S}^{l}[t]$. Moreover, it contains four learnable parameters $\mu _{D}^{l}$,$\mu _{S}^{l}$,$\lambda _{D}^{l}$ and $\lambda _{S}^{l}$. The LM-H model significantly enhances the representation ability and adaptability by regulating the dual membrane potentials and multiple learnable parameters. Following the similar idea of STC-LIF and STC-PLIF, $\boldsymbol \gamma ^{l}[t]$ regulates the input $\boldsymbol{x}^{l}[t]$, while $\boldsymbol \beta ^{l}[t]$ regulates the key membrane potential $\boldsymbol{v}_{S}^{l}[t]$. In addition, in the soft reset process of the LM-H model, the gradient backpropagation of $\boldsymbol{s}^{l}[t]$ is truncated. We find that this increases the gradient calculation error, so we choose not to truncate the gradient information of $\boldsymbol{s}^{l}[t]$.

\begin{figure*}[h]
\vskip 0.2in
\begin{center}
\centerline{\includegraphics[width=\linewidth]{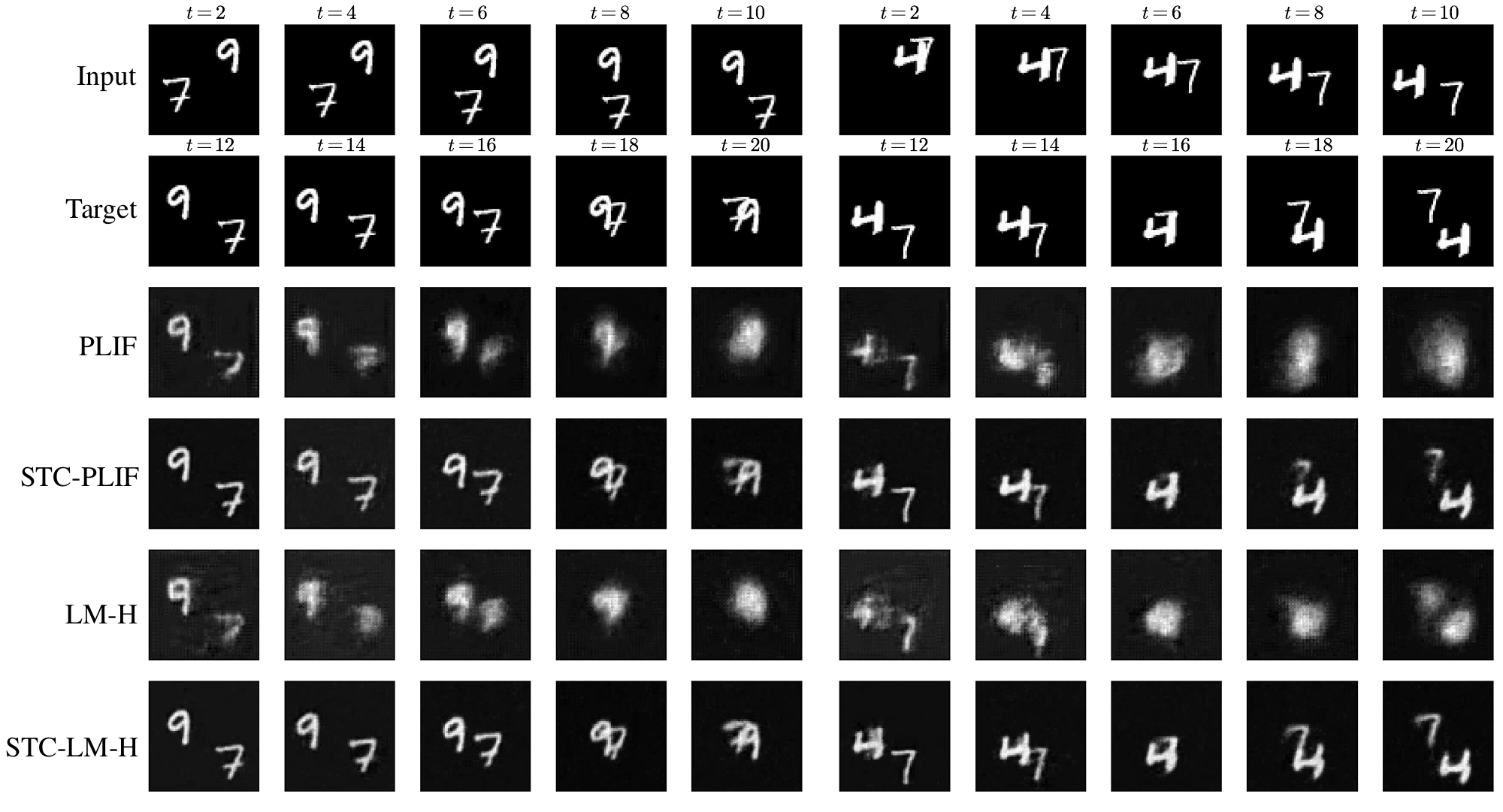}}
\caption{Comparison of the qualitative visualization prediction results of original models and enhanced models on the Moving MNIST dataset.}
\label{p6}
\end{center}
\vskip -0.2in
\end{figure*}

\textbf{Analysis of Enhanced Models.} For the enhanced models, it is obvious that they have more complex dynamic characteristics than the original models. When $\boldsymbol \beta ^{l}[t]$ and $\boldsymbol \gamma ^{l}[t]$ equals 0, the enhanced models are equivalent to the original models, Therefore, the spatio-temporal circuit is an improvement in the basic representation ability of the spiking neuron models. The comparison of the qualitative visualization results of the enhanced models and the original models on Moving MNIST is shown in Figure~\ref{p6}, which shows that the spatio-temporal circuit significantly improves the models’ performance in capturing deep spatio-temporal information features and modeling spatio-temporal dependencie.


\section{Additional Experiments}
\textbf{Experiments on static classification tasks.} We have conducted experiments on the CIFAR100 dataset using the VGGSNN (64C3-128C3-MP2-256C3-256C3-MP2-512C3-512C3-MP2-512C3-512C3-MP2-FC) architecture. The results are presented in Table~\ref{table10}. It is evident that the STC-LIF model is also beneficial in improving the performance of classification tasks. The accuracy of the STC-LIF model surpasses that of the vanilla LIF model at the same time step. Notably, this advantage becomes more pronounced as the time step increases. Consequently, we believe that spiking neural networks, with their remarkable spatial and temporal processing capabilities, possess substantial potential for advancements in the domain of complex dynamic task processing. 

\begin{table}[htbp]
\caption{Comparsion of different methods on the CIFAR100 dataset.}
\centering
\vskip 0.1in
\resizebox{0.30\columnwidth}{!}{
\begin{tabular}{ccc}
   \toprule

   Method&Time-steps&Accuracy(\%) \\ 
   \midrule
    Vanilla LIF&4&63.40\\        
    STC-LIF&4&66.64\\        
    Vanilla LIF&6&63.99\\        
    STC-LIF&6&68.93\\  

   \bottomrule
\end{tabular}
}
\label{table10}
\vskip -0.1in
\end{table}

\textbf{Experiments with different self-connection methods.} We have conducted more comprehensive experiments on the Moving MNIST dataset, and the corresponding results are presented in Table~\ref{table11}. It has been observed that employing self-connections alone can effectively enhance the dynamic representation of spiking neurons, resulting in a significant improvement compared to the vanilla LIF model. However, this approach has certain drawbacks. Firstly, it lacks flexibility and limits the representational capacity of spiking neurons, thereby leading to inferior performance compared to the STC-LIF model with group convolution or global convolution. Secondly, it entails a higher memory cost due to the assignment of weight parameters to each neuron. While global convolution yields the best performance, it comes at the expense of substantial computational requirements. Therefore, considering the overall trade-off, utilizing group convolution emerges as a favorable choice. It significantly enhances the model's representational capacity without imposing a significant increase in computational cost.

\begin{table}[htbp]
\caption{Comparison of models with self-connection, group convolution and global convolution on the Moving MNIST dataset.}
\centering
\vskip 0.1in
\resizebox{0.6\columnwidth}{!}{
\begin{tabular}{ccccc}
   \toprule
   Model&Operation&MSE↓&MAE↓&Parameters(M) \\ 
   \midrule
    Vanilla LIF&-&102.8&246.2&3.305\\        
    STC-LIF&Only self-connection&64.5&177.6&6.451\\        
    STC-LIF&Group convolution(groups=16)&47.0&136.4&3.922\\    
    STC-LIF&Global convolution&38.5&108.1&13.138\\  

   \bottomrule
\end{tabular}
}
\label{table11}
\vskip -0.1in
\end{table}

\end{document}